\documentclass[runningheads]{llncs}

\usepackage[T1]{fontenc}
\usepackage{amsmath,amssymb}
\usepackage{booktabs}
\usepackage{adjustbox}
\usepackage{wrapfig}
\usepackage{graphicx}
\usepackage{alltt}
\usepackage[table]{xcolor}
\usepackage{simpleicons}
\usepackage[most]{tcolorbox}
\usetikzlibrary{arrows.meta}

\newcommand{\llmicon}[1]{%
  \raisebox{-0.2ex}{\includegraphics[height=1.6ex]{figures/icons/#1.pdf}}%
  \hspace{0.45em}%
}

\newcommand{\pholm}{p_{\mathrm{Holm}}}
\newcommand{\bench}{CovenantBench}

\usepackage{fontawesome5}
\usepackage{hyperref}
\usepackage{cleveref}
\definecolor{oursbg}{HTML}{D9E6F5}
\crefname{table}{Table}{Tables}\Crefname{table}{Table}{Tables}
\crefname{figure}{Fig.}{Figs.}\Crefname{figure}{Fig.}{Figs.}
\crefname{section}{Section}{Sections}\Crefname{section}{Section}{Sections}
\definecolor{cdkw}{HTML}{0000FF}    
\definecolor{cdctl}{HTML}{AF00DB}    
\definecolor{cdfn}{HTML}{795E26}     
\definecolor{cdtype}{HTML}{267F99}  
\definecolor{cdvar}{HTML}{001080}    
\definecolor{cdnum}{HTML}{098658}   
\definecolor{cddate}{HTML}{A31515} 
\definecolor{cdbg}{HTML}{F7F7F7}     
\definecolor{cdtext}{HTML}{EEF3FA}   
\definecolor{cdedge}{HTML}{C8C8C8}

\graphicspath{{figures/}{figures/icons/}}
\begin{document}

\title{FormaFin: Verifiable Financial Analysis with Solver-Guided Reinforcement Learning}
\titlerunning{FormaFin: Verifiable Financial Analysis}

\author{Bethel Hall \and William Eiers}
\authorrunning{B. Hall and W. Eiers}

\institute{
Stevens Institute of Technology, Hoboken, NJ, USA \\[8pt]
\faGithub\ \href{https://github.com/verifiedfinance/loan-agent}{Code}
\qquad
\href{https://huggingface.co/bethelhall/formafin}{%
\raisebox{-0.9ex}{\includegraphics[height=3.5ex]{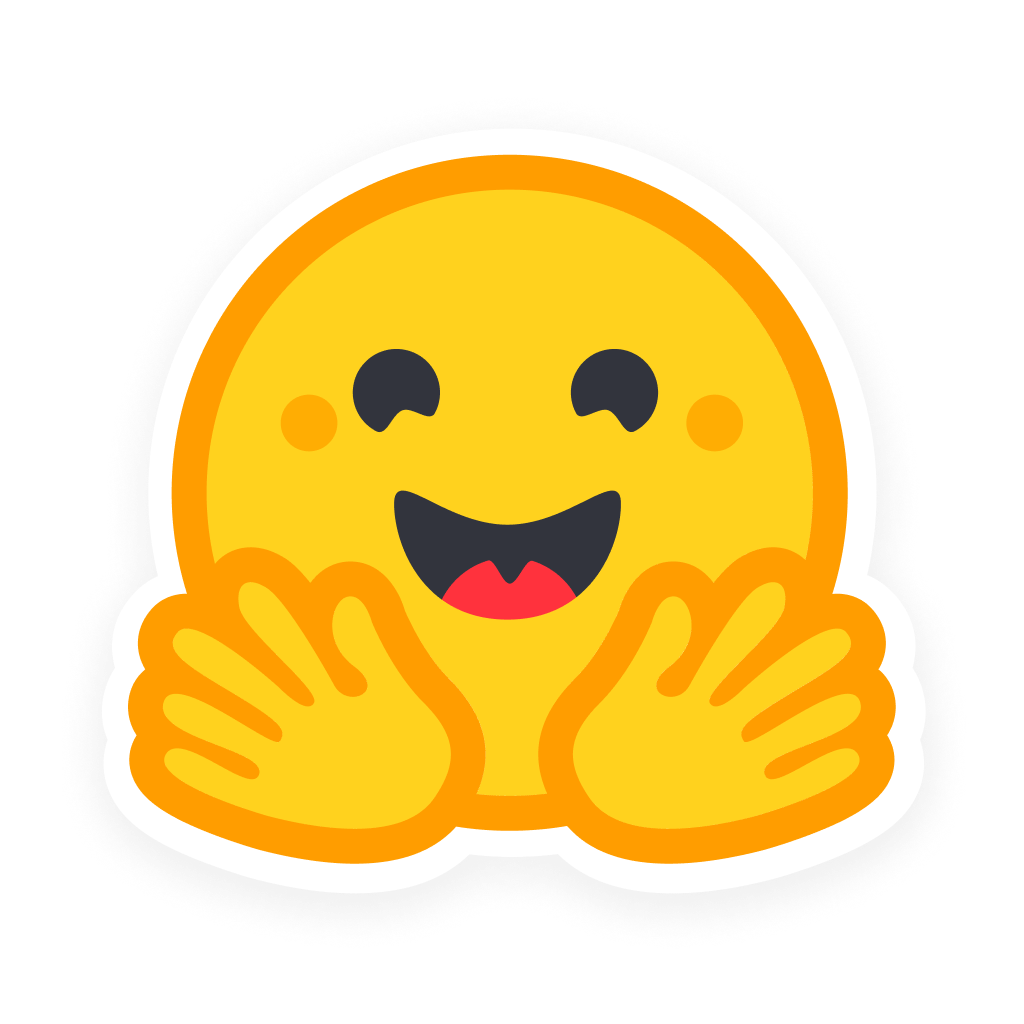}}\ Model%
}
}
\maketitle

\begin{abstract}

Formally verifying financial analyses generated by large language models (LLMs) remains challenging as financial rules are largely expressed in natural language and rarely have corresponding formal specifications. Training language models to formalize these rules is hindered by the scarcity of labeled data and reliable training signals, while confidential financial data favor small models that run locally. We introduce FormaFin, a framework that trains small open language models to translate financial rules into reusable SMT formulas. FormaFin integrates teacher distillation with reinforcement learning using an SMT solver to compute rewards from numerical thresholds stated in financial rules. We evaluate FormaFin on corporate loan monitoring, in which lenders check borrowers' financial figures against loan agreements. To evaluate our framework, we curate \bench{}, a benchmark containing 143 loan agreements and 2,527 questions covering compliance, earnings cushion, and debt capacity. Our results show that a 7B formalizer achieves 94.8\% accuracy, outperforming open LLMs up to ten times larger. Across three model families, accuracy ranges from 92.6 to 96.8\%. Our framework also detects 95.6 to 99.7\% of errors in other LLMs' financial analyses. We make FormaFin and \bench{} publicly available.

\end{abstract}


\begin{figure}[t]
\centering
    \includegraphics[width=\textwidth]{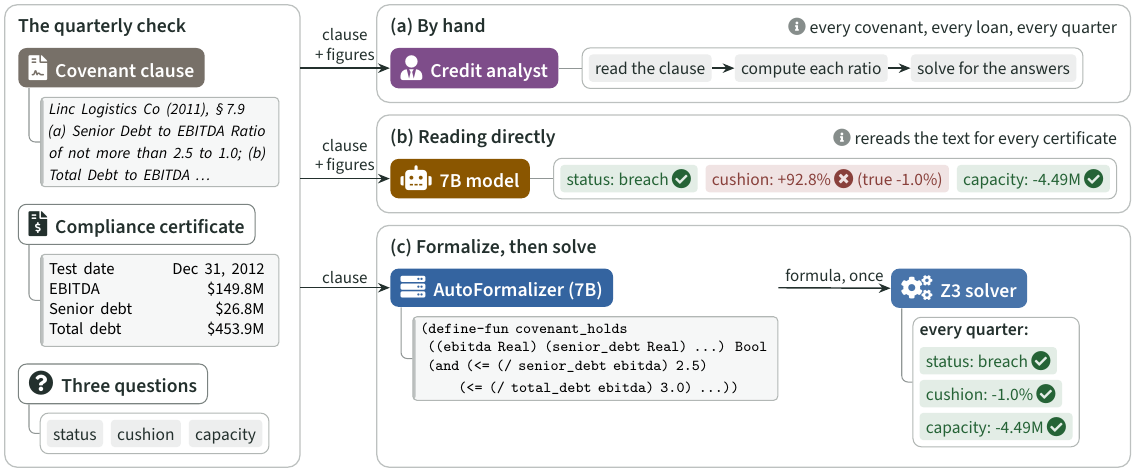}
\caption{The quarterly covenant check using a real loan  covenant case, and a constructed compliance certificate showing a 1\% breach. (a) An analyst checks compliance against loan covenant. (b) A 7B model checking the clause gets the earnings cushion wrong by 94 points. (c) Trained with
FormaFin, the same model writes the covenant once as a formula, and the Z3 solver computes all three answers exactly,
every quarter.}

\label{fig:overview}
\end{figure}

\section{Introduction}
\label{sec:intro}

Artificial intelligence is increasingly used in financial services. A 2026 Bank of England and Financial Conduct
Authority survey found that 86\% of responding firms use AI \cite{boefca2026ai}. Financial institutions routinely check reported figures against contractual, regulatory, and internal rules. Lenders check whether borrowers meet loan covenants
(financial conditions they must satisfy under their loan
agreements), banks monitor exposure limits, and asset
managers check whether portfolios comply with investment
mandates. These checks recur as new figures arrive, carry financial and legal
consequences, and often involve confidential data.

We focus on corporate lending, where loan agreements specify financial conditions that borrowers must satisfy. Each reporting period, lenders compare borrowers' reported figures against these conditions to determine compliance and how much financial headroom remains before a breach. Covenant monitoring therefore provides a concrete setting for studying the reliability of LLM-based financial analysis. LLMs offer a way to automate these
repeated checks, but a correct answer requires interpreting
the covenant, identifying the applicable limits, and
calculating compliance and headroom. When an LLM performs
these steps directly, verifying its answer may require
repeating the same analysis. The difficulty is particularly
apparent in breach cases, where the model must identify the
violation and calculate how far the reported figures must
change to restore compliance. On our benchmark, an
off-the-shelf 7B model that reads the clause and certificate and answers directly, which we call a \emph{reader},
answers 76.5\% of questions correctly with 20\% headroom,
but only 44.7\% when the borrower is 1 to 5\% beyond a
limit. A 70B model scores 94.6\% and 52.2\%, respectively.
\textbf{We hypothesize that LLMs can produce more reliable
financial analyses by translating financial rules into
machine-checkable representations rather than making each
decision directly.} An LLM formalizes each rule once, and
a solver applies it to new figures, computes headroom,
and checks subsequent analyses (\cref{fig:overview}).

This approach builds on autoformalization in mathematics, where LLMs translate informal statements into 
formal representations that can be checked by a theorem prover~\cite{wu2022autoformalization,huang2025formarl}. Existing approaches use LLMs to generate executable programs or formal specifications that external tools can evaluate \cite{gao2023pal,ye2023satlm,pan2023logiclm,hall2026verifin}. However, these approaches typically generate a new representation for each question. Financial institutions also face confidentiality constraints that can restrict externally hosted models.
Small open models offer local deployment, control over model versions for validation \cite{sr117}, and the ability to
customize training. However, training these models to formalize financial rules is challenging. Mathematical autoformalization benefits from established formal languages such as Lean and large libraries of formalized theorems. In contrast, financial agreements rarely come with equivalent machine-checkable specifications or paired examples for training. Without such references, evaluating generated formulas and defining reliable training rewards are difficult.

We address these challenges with FormaFin (\cref{fig:method}). Our key insight is that numerical thresholds specified in financial rules provide a training signal because they determine when a requirement is satisfied or violated. FormaFin first distills formulas from a teacher model. It then uses reinforcement learning with solver-computed rewards that test generated formulas at the stated limits across dates and conditions, without requiring reference formulas for the reward. For deployment, Z3 checks whether two separately trained formalizers produce equivalent formulas. We use agreed formulas for automated checks and refer disagreements to an analyst.

We make the following contributions:
\begin{itemize}

\item \textbf{A training recipe for financial autoformalization.} We train small open models without labels,
using teacher distillation and reinforcement learning with a reward computed by the solver, followed by a proof of
equivalence that selects formulas for automated use (\cref{sec:method}).

\item \textbf{CovenantBench.} We construct, to our knowledge, the first benchmark for covenant compliance with 143 covenants from SEC credit agreements whose borrowers never appear in training and 1,064 compliance
certificates (\cref{sec:benchmark}).

\item \textbf{Empirical evidence.} Our formalizers with 7 to 8 billion parameters reach 92.6 to 96.8\% accuracy
across three model families, higher than open readers up to ten times larger that read the clause and check compliance
themselves. Rewards from solver tests outperform an LLM judge as the training reward (94.8\% vs 91.4\% accuracy).
When we use the formulas to check the analyses of other LLMs, they catch 98 to 99.7\% of their mistakes. Proofs also
find mistakes that testing misses. One in eight formulas that are correct on every benchmark certificate is still
wrong on other certificates (\cref{sec:experiments}).

\end{itemize}

\section{Related Work}
\label{sec:related}

\textbf{LLMs for finance.} BloombergGPT \cite{wu2023bloomberggpt} and FinGPT \cite{yang2023fingpt}
adapt language models to financial text; FinQA, TAT-QA, CUAD and LegalBench
\cite{chen2021finqa,zhu2021tatqa,hendrycks2021cuad,guha2023legalbench} evaluate numerical
reasoning over reports and contract understanding. All evaluate a model that reads a document and
answers a question. Covenant violations are frequent and transfer control rights to lenders
\cite{dichev2002covenant,chava2008covenants}, yet no benchmark covers the inverse computations a
lender performs every quarter.

\textbf{Solver-aided LLMs.} PAL generates Python \cite{gao2023pal}; SatLM and Logic-LM generate
declarative specifications \cite{ye2023satlm,pan2023logiclm}. All prompt a large model per
question. In finance, VeriFin \cite{hall2026verifin} verifies LLM claims against XBRL filings with
Z3, Hsia et al.\ \cite{hsia2026neurosymbolic} pair an LLM with an SMT solver for regulatory
enforcement cases, and AuditFlow \cite{wang2026auditflow} checks XBRL consistency with
tool-calling agents. VeriMed \cite{hall2026neurosymbolic} treats solver-proven \emph{disagreement}
among sampled formalizations of a medical-device requirement as a sign of ambiguity; Stage III
below instead uses proven \emph{agreement} between separately trained models to decide which
formulas run unreviewed. None of these trains the model to write a rule once, so that the same
formula verifies every later analysis.

\textbf{Autoformalization and verifiable rewards.} Autoformalization targets mathematics
\cite{wu2022autoformalization}. Stratified Consistency Distillation \cite{hou2026scd} finetunes a
7B model on frontier-LLM SMT translations that several samples agree on, and FormaRL
\cite{huang2025formarl} rewards statements that compile and that an LLM judges faithful.
Verifiable-reward RL normally compares outputs to known answers
\cite{lambert2024tulu3,deepseek2025r1}, often with GRPO \cite{shao2024deepseekmath}. Financial
rules have no known answers but state their own limits, which we turn into solver tests, requiring
neither reference formulas nor an LLM judge.

\section{Problem Setting}
\label{sec:prelim}

\textbf{The task.} Given a loan covenant and a compliance certificate containing reported figures, a reporting date, and applicable conditions, the task is to produce three answers. The \emph{compliance status} is ``comply'', ``breach'', or ``undetermined'' when required figures are missing. The \emph{earnings cushion} is the percentage by which earnings can fall before a breach. It is negative when the borrower is already in breach and ``none'' if no level of earnings would cause a breach. The \emph{debt capacity} is the dollar amount by which debt can increase before a breach, or ``none'' if no level of debt would cause a breach.

\textbf{Formalization.} We use satisfiability modulo theories (SMT), which extends Boolean satisfiability with
arithmetic, so a single formula can combine logical conditions with numerical limits. A language model translates
each clause once into a formula in SMT-LIB, the standard input language of SMT solvers \cite{barrett2010smt}. The
formula has a fixed parameter list, with one real number per reported figure, the reporting date as a YYYYMMDD
integer, and one Boolean for each condition that changes a limit. We evaluate the formula with the SMT solver Z3
\cite{demoura2008z3}.

Given a certificate, the \textit{status} is comply if the formula must hold, breach if it cannot hold, and
undetermined if missing figures allow both. A search over the earnings or debt figure then finds the
\textit{cushion} and the \textit{capacity}, the points at which the status changes. \textbf{Verification.} Let $\varphi$ and $\psi$ be two formulas for the same clause and $c=(x,d,b)$ a certificate
with figures $x$, reporting date $d$ and conditions $b$. The formulas are \emph{equivalent} if $\varphi(c)=\psi(c)$ for
every certificate with positive figures and no upper bound, positive denominators in every reference ratio, a
reporting date on any calendar day from 1990 to 2030, and any combination of events. Z3 decides equivalence by
searching for such a certificate with $\varphi(c)\neq\psi(c)$. If none exists, the two formulas agree on every
such certificate, every date in between included. If one exists, the solver returns it as a counterexample. A
formula we trust (\cref{sec:method}, Stage III) also verifies any analysis of the same clause. We accept an answer
to $c$ if it matches the answer computed from $\varphi(c)$ and flag it otherwise.

\section{Methodology}
\label{sec:method}

\begin{figure}[t]
\centering
\includegraphics[width=0.88\textwidth]{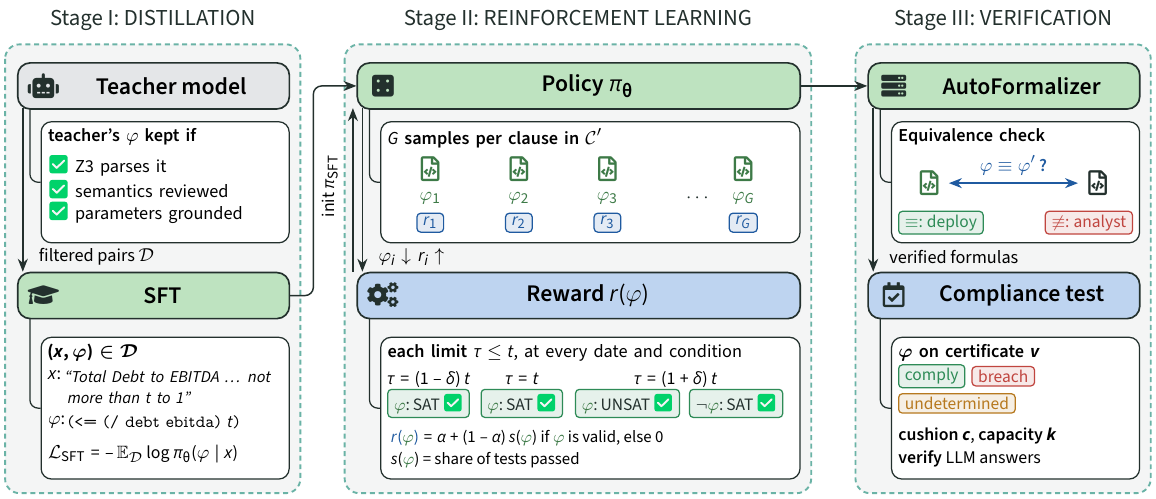}
\caption{How FormaFin works. \textbf{Stage~I.} Teacher formulas that pass automatic checks train the small model.
\textbf{Stage~II.} The solver tests each sampled formula just inside, at and just outside every stated limit, and the
reward grows with the share of tests passed. \textbf{Stage~III.} Formulas that a second formalizer provably matches are
used without review, and the rest go to an analyst.}
\label{fig:method}
\end{figure}

Our framework has three stages for training a small open model to translate a covenant clause into the formula
defined in \cref{sec:prelim}, without requiring ground-truth annotations (\cref{fig:method}). \textbf{Stage I} distills formulas from a
teacher model. \textbf{Stage II} refines the model with reinforcement learning, using a reward that the solver computes from
the limits stated in each clause. In \textbf{Stage III}, we accept a formula only when the solver proves it equivalent to the
formula of a second, independently trained model. We instantiate the recipe for loan covenants. Our training approach requires rules that state their limits in text, figures reported against those
rules, a teacher model and a solver. Regulatory ratios, concentration limits and fund mandates satisfy these
requirements, and we discuss them in \cref{sec:discussion}.

\textbf{Stage I: teacher distillation.}
In this stage, we ask a teacher model to translate 1,000 training clauses into SMT formulas, with comments linking each parameter to the original clause. We filter the formulas in three steps. First, Z3 checks whether each formula parses. Second, we check that every parameter is grounded in the clause. Third, a separate model checks whether the formula accurately represents the clause. 

After filtering, 789 pairs remain for Supervised Fine-tuning (SFT) of a small model with a LoRA adapter \cite{hu2022lora}, using sequence-level knowledge distillation \cite{kim2016sequence}.

\textbf{Stage II: reinforcement learning from solver feedback.} Stage II trains on clauses without reference formulas
or answers. Our key observation is that covenants state their own limits, and these limits tell us how a correct
formula must behave. A teacher model lists each clause's financial items, limits, dates and conditions, which tell us
where to test. We keep the 828 clauses whose thresholds and dates all appear in the text, 242 of them with schedules
or conditions. The model being trained sees only the clause and its parameter list.

\emph{Reward.} For each limit, Z3 tests the generated formula $\varphi$ with the limited ratio fixed at three values.
For ``Total Debt to EBITDA of not more than 3.0 to 1.0'', a borrower with a ratio of 2.94 or 3.0 must be able to
comply, so $\varphi$ must be satisfiable there. A borrower with a ratio of 3.06 must be in breach, so $\varphi$ must be
unsatisfiable while $\neg\varphi$ is satisfiable, which stops a formula that rejects every certificate from passing.
In general we test at $(1-\delta)t$, $t$ and $(1+\delta)t$ for a limit $t$, reverse the tests for lower limits, and
repeat them at each date and condition where the limit applies. With $s(\varphi)$ the share of tests passed,
\[
r(\varphi)=\begin{cases}
0 & \text{if $\varphi$ does not parse or violates the parameter list,}\\
\alpha+(1-\alpha)\,s(\varphi) & \text{otherwise,}
\end{cases}
\]
where $\alpha=0.3$ gives any valid formula more reward than an invalid one and $\delta=2\%$. Both are fixed before
training. 
\emph{Training.} We optimize this reward with GRPO \cite{shao2024deepseekmath}. For each clause, the model samples
$G=4$ formulas, and each formula's advantage is its reward normalized within the group,
$\hat{A}_i=(r_i-\operatorname{mean}(r))/\operatorname{std}(r)$. Passing the tests does not show that a formula uses
the right financial items, so Stage III checks agreement between separately trained formalizers.

\textbf{Stage III: selecting formulas without labels.}
During evaluation, Z3 can prove a formula equivalent to a reference over the entire input domain. In practice, no reference exists, so we compare formulas from two separately trained models. If Z3 proves them equivalent, we use the formula to check compliance and verify later LLM analyses. Otherwise, we refer the covenant to an analyst with a counterexample showing where the formulas differ. Because different seeds share the Stage I adapter and may make the same errors, we also compare models from different families (\cref{sec:verification}).

\section{Experimental Setup}

\label{sec:setupsec}

\subsection{\bench{}}
\label{sec:benchmark}

\textbf{Covenants.} We extract 12,517 financial-ratio clauses from 14,526 credit agreements filed with the SEC \cite{ganglmair2026complexity}. Because companies reuse their own wording, splitting by agreement would leave 64\% of test clauses from borrowers seen in training.  We therefore use only unseen borrowers for testing and exclude test clauses whose token sets have a Jaccard similarity of 0.7 or more to any training clause. From the training borrowers we draw two disjoint sets of 1,000 clauses, one for Stage I and one for Stage II.

\textbf{References.} We use Claude Sonnet 5.5 and Qwen3.8-27B to independently translate each test clause into a specification of its limits. We keep only clauses for which the specifications agree exactly, and every threshold appears in the original text. We encode the agreed specifications as SMT formulae for the equivalence checks in RQ4. This produces 143 clauses from 108 agreements dated 1993--2015, covering coverage, capitalization, leverage, liquidity, and asset related covenants. Among the 143 covenants, 32 have multiple limits, and 27 include date schedules or conditions.

\textbf{Certificates and scoring.} We construct compliance certificates by choosing one reporting date per schedule step and adjusting earnings, or debt when earnings is absent, to produce headroom of $-5\%$, $-1\%$, $+1\%$, $+5\%$, or $+20\%$. We also add cases with missing figures and, where applicable, changed conditions, yielding 1,064 certificates. For each certificate, we use the agreed specification to compute the reference compliance status, earnings cushion, and debt capacity. Each certificate asks for compliance status, earnings cushion if the clause contains an earnings item, and debt capacity if it contains a debt item, giving 2,527 questions. Status requires an exact match, cushion is correct within 0.5 percentage points, and capacity within 0.5\% of current debt. We average scores within each clause and then across clauses.

\subsection{Baselines and Training Setup}
\label{sec:setup}

\textbf{Baselines and evaluation.}
We compare FormaFin with readers, open models that receive a covenant clause and compliance certificate and answer the three questions directly, with up to 4,096 reasoning tokens. We evaluate the base model of our formalizer, Qwen2.5-Coder-7B-Instruct, and three larger models, Qwen3.8-27B, Qwen2.5-Coder-32B, and Llama-3.3-70B. All models run locally, with the 70B model in 4-bit and the others in bf16. We evaluate FormaFin after each training stage, using Z3 to compute its answers, and repeat the training recipe with Llama-3.1-8B and Mistral-7B. We compare results using two-sided sign-flip tests over clauses (10,000 permutations), 95\% clause-bootstrap intervals, and Holm correction.

To evaluate our reward design, we train the 7B model with two alternative Stage II rewards. \emph{Binary} uses the same solver tests but assigns 1 only if all tests pass. \emph{LLM judge} adapts the FormaRL reward \cite{huang2025formarl}, assigning 1 if the formula passes the parameter list check and Claude Sonnet 5.5 judges it consistent with the clause, and 0 otherwise.

\textbf{Training details.}
In Stage I, we train a LoRA adapter (rank 16, scaling 32, all linear layers) for one epoch at a learning rate of $10^{-4}$. In Stage II, we continue training with GRPO, sampling four formulae per clause at temperature 0.8 and top-$p$ 0.95, with a maximum of 1,024 tokens. We use a learning rate of $5\times10^{-6}$ with linear decay, clipping at 0.2 and no KL penalty. We train for 1,000 steps, saving a checkpoint every 250 steps, and run three seeds per model family on a 96\,GB GPU. \\

\section{Results}
\label{sec:experiments}
We organize the results around four research questions. \textbf{RQ1}~Does a small formalizer outperform larger readers(\S\ref{sec:main})? \textbf{RQ2}~Does the training recipe generalize across model families(\S\ref{sec:families})? \textbf{RQ3}~Can its formulas catch the errors in other LLMs' analyses
(\S\ref{sec:llmverify})? \textbf{RQ4}~Can we identify reliable formulas without ground-truth labels(\S\ref{sec:verification})?

\subsection{RQ1: A small formalizer outperforms readers up to ten times larger}
\label{sec:main}

\begin{table}[t]
\centering
\footnotesize
\definecolor{oursbg}{HTML}{EAF4EC}
\setlength{\tabcolsep}{5pt}
\setlength{\aboverulesep}{0pt}\setlength{\belowrulesep}{0pt}
\renewcommand{\arraystretch}{1.2}
\newcommand{\pmsd}[1]{\,{\scriptsize\textcolor{gray}{$\pm$#1}}}
\newcommand{\nosd}{\phantom{\pmsd{0.0}}}
\caption{Accuracy on \bench{} (\%), overall and by question (status, earnings cushion, debt capacity). Proven -- formulas
proven equivalent to the reference. $\pm$: standard deviation over three training runs.}
\label{tab:main}
\begin{tabular}{lccccc}
\toprule
 & Accuracy & Status & Cushion & Capacity & Proven \\
\midrule
\multicolumn{6}{l}{\textit{Readers}} \\
\llmicon{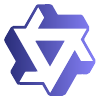}Qwen2.5-Coder-7B      & 64.8\nosd & 86.6 & 34.1 & 60.4 & -- \\
\llmicon{qwen}Qwen3.8-27B           & 88.8\nosd & 89.5 & 87.5 & 84.3 & -- \\
\llmicon{qwen}Qwen2.5-Coder-32B     & 85.8\nosd & 92.4 & 75.5 & 82.9 & -- \\
\llmicon{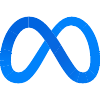}Llama-3.3-70B (4-bit) & 79.1\nosd & 91.6 & 67.9 & 71.0 & -- \\
\midrule
\multicolumn{6}{l}{\textit{FormaFin (formalize, then solve)}} \\
\rowcolor{oursbg}\llmicon{qwen}Qwen2.5-Coder-7B & 94.8\pmsd{1.0} & 95.7 & 95.0 & 93.5 & \textbf{80.9} \\
\rowcolor{oursbg}\llmicon{meta}Llama-3.1-8B     & \textbf{96.8}\pmsd{1.1} & \textbf{97.9} & \textbf{95.8} & \textbf{96.7} & 78.1 \\
\rowcolor{oursbg}\llmicon{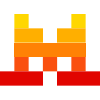}Mistral-7B    & 92.6\pmsd{1.3} & 95.3 & 89.3 & 93.3 & 73.2 \\
\bottomrule
\end{tabular}
\end{table}

FormaFin outperforms every reader (\cref{tab:main}). Our 7B formalizer scores $94.8\pm1.0\%$: 6.1 points above
Qwen3.8-27B ($\pholm=0.004$), 9.0 above Qwen2.5-Coder-32B, 15.7 above Llama-3.3-70B, and 30.0 above its own base model as a reader
(the remaining comparisons have $\pholm<0.001$). The advantage holds for every seed. Size does not close the gap: the 70B reader
scores 79.1\%, below both mid-sized readers and 13 to 18 points below every FormaFin model.

We see the largest gains on questions about financial headroom. On the earnings cushion, the same 7B model is correct 34.1\% of the time when it reads the clause and 95.0\% when it
formalizes it. The reader achieves 86.6\% accuracy on compliance status that requires only checking whether the limits are met. The larger gain in earnings cushion suggests that formalization is particularly useful for questions requiring numerical reasoning beyond a compliance check.

\textbf{Solver-guided reinforcement learning improves on teacher distillation.}
Without training, the formalizer scores 53.2\%, below the same model as a reader, with 61 unusable formulas. Stage I improves accuracy to 85.8\%. We start with 1,000 training clauses. Z3 rejects 48 formulas that fail to parse, the grounding check rejects 15 with undeclared parameters, and a separate model rejects 123 judged inconsistent with the clause. Another 25 produces unusable pair because the teacher produces no formula or the pair exceeds the training context, leaving 789 pairs for Stage I. Stage II then adds 9.0 points ($\pholm<0.001$), improving all three questions (status $+8.5$, cushion $+13.4$, capacity $+8.9$, all $\pholm\le0.004$) and raising the share of formulas proven equivalent to the reference from 67\% to 81\%.

\begin{figure}[t]
\centering
\definecolor{rwours}{HTML}{1B8A3A}\definecolor{rwjudge}{HTML}{D62728}\definecolor{rwbin}{HTML}{7B3294}
\definecolor{rwink}{HTML}{262626}\definecolor{rwgray}{HTML}{8C8C8C}
\providecommand{\rwkey}[3]{\tikz[baseline=-0.55ex, x=1cm, y=1cm]{\draw[#1, line width=0.9pt, #2] (0,0)--(0.46,0); #3}}
\resizebox{\textwidth}{!}{%
\begin{tikzpicture}[x=1pt, y=1pt, font=\fontfamily{phv}\selectfont]
\node[anchor=south west, inner sep=0] at (0,0)
  {\includegraphics{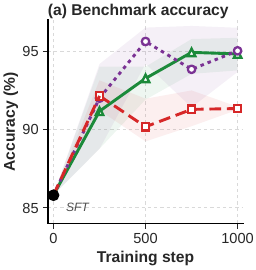}};
\draw[rwgray, line width=0.6pt] (120,106.6) -- (122.5,106.6) -- (122.5,76.6) -- (120,76.6);
\node[anchor=west, inner sep=0, text=rwink] (T) at (134,84) {%
  \fontsize{6.6}{8}\selectfont\setlength{\tabcolsep}{3pt}\renewcommand{\arraystretch}{1.4}%
  \begin{tabular}{@{}l@{\hspace{5pt}}rr@{\hspace{4pt}}}
  \toprule
  \textbf{Reward} & \textbf{Acc.} & \textbf{Proven} \\
  \midrule
  \rwkey{rwours}{}{\draw[rwours, fill=white, line width=0.7pt] (0.23,0.075)--(0.165,-0.045)--(0.295,-0.045)--cycle;}\,FormaFin & 94.8 & 80.9 \\
  \rwkey{rwjudge}{dash pattern=on 3.6pt off 1.8pt}{\draw[rwjudge, fill=white, line width=0.7pt] (0.175,-0.055) rectangle (0.285,0.055);}\,LLM judge & 91.4\rlap{\textsuperscript{**}} & 76.0\rlap{\textsuperscript{**}} \\
  \rwkey{rwbin}{dash pattern=on 1.1pt off 1.3pt}{\draw[rwbin, fill=white, line width=0.7pt] (0.23,0) circle (0.055);}\,Binary & 95.0 & 81.1 \\
  \bottomrule
  \end{tabular}};
\node[anchor=base west, inner sep=0, text=rwink, font=\fontfamily{phv}\selectfont\bfseries\fontsize{7.5}{9}\selectfont]
  at (T.west |- 0,121.6) {(b) Final result};
\node[anchor=north, inner sep=0, text=rwgray, font=\fontfamily{phv}\selectfont\fontsize{6}{7}\selectfont]
  at ([yshift=-4pt]T.south) {at step 1000};
\draw[rwgray, line width=0.6pt, -{Stealth[length=3.2pt]}] (122.5,91.6) -- (T.west);
\node[anchor=south west, inner sep=0] at ([xshift=8pt]T.east |- 0,0)
  {\includegraphics{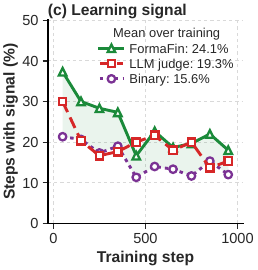}};
\end{tikzpicture}}
\caption{Solver tests outperform an LLM judge as the Stage II reward. (a) Benchmark accuracy during training. (b)
Final accuracy and share of formulas proven equivalent to the reference, in \% ($^{**}$\,below FormaFin,
$\pholm<0.01$). (c) Share of training steps with a learning signal.}
\label{fig:rewards}
\end{figure}

\textbf{Solver rewards outperform an LLM judge.}
We replace the Stage II reward with two alternatives 
and keep the model, data, clause order, steps and seeds fixed to avoid confounds (\cref{fig:rewards}). With an LLM
judge as the reward, the final model is 3.5 points less accurate and has 4.9 points fewer formulas proven equivalent
to the reference ($\pholm\le0.005$). All three rewards improve the model equally for the first 250 steps. After
that, the judge stalls near 91\%, while both rewards computed by the solver reach 95\%. The remaining errors are
small, such as a wrong threshold, and only running the formula at the limit reveals them. In covenant monitoring
these errors matter most, because a threshold or its strictness decides whether a borrower is in breach. Giving
credit only when every test passes reaches the same final result as partial credit (95.0 against 94.8\% accuracy).
Partial credit lets the model learn from more training steps (24.1 against 15.6\%), but this does not change where
training ends.

\textbf{Does the label-free reward track accuracy?} We relate each checkpoint's benchmark score to the mean training
reward over the 100 steps before it. Across all checkpoints, the two correlate strongly
(Pearson $r=0.88$ over 45 checkpoints of three seeds and three families): when the reward rises, the model answers
more benchmark questions correctly, although the reward never sees a benchmark answer. Among the RL checkpoints alone,
which all score between 88.5 and 97.8\%, the correlation is weak ($r=0.40$), because the reward saturates near 0.93.
The reward therefore signals large improvements reliably, but it does not rank strong models finely; for that, we rely
on the solver's equivalence checks (\cref{sec:verification}).


\textbf{Errors depend on the figures.} Readers are least accurate on breaches, where a lender depends on them most.
Compared with cases where earnings could fall 20\% before a breach, the 7B and 70B readers lose 31.8 and 42.4 points
on breaches ($\pholm=0.001$), the 32B reader loses 8.4, and the 27B reader's gap of 4.1 is not significant. The 7B and
70B readers still report the breach in 75.9 and 83.2\% of these cases but give the correct negative cushion in only
4.6 and 27.5\%. FormaFin scores 94.9, 94.6 and 94.8\% on breaches, thin margins and ample headroom, because it
interprets each clause once and a correct formula answers every certificate correctly.

\subsection{RQ2: Our training recipe transfers across model families}
\label{sec:families}

We apply the same training recipe to Llama-3.1-8B and Mistral-7B, two general purpose models, and observe the same pattern
(\cref{tab:main}). The three families start far apart, from 25.0 to 64.8\% as readers, and from 34.9 to
85.8\% after SFT, and end close together, at 92.6 to 96.8\%. We see the largest improvement with Llama-3.1-8B, which starts with the lowest Stage I accuracy but gains 62.0 points after reinforcement learning and achieves the highest final accuracy. Its SFT formulas fail mostly on
syntax rather than logic. Stage I produces 82 unusable formulas out of 143. To further understand this, we manually inspected the failure modes and observed that 56 have malformed parameter lists, usually missing parentheses required by SMT-LIB, 21 exceed the 1,024-token limit, and 5 contain other syntax errors. After reinforcement learning, only one remains unusable. Every
family scores above every reader. Mistral-7B scores 92.6\% against 88.8\%
for Qwen3.8-27B, the strongest baseline, although the difference is not statistically significant ($p=0.10$).

\subsection{RQ3: FormaFin verifies LLM analyses}
\label{sec:llmverify}

Note that the goal of FormaFin is to verify financial analyses, not only to produce answers. We use formulas generated by our 7B formalizer to check the answers of each reader. (\cref{tab:llmverify}). We accept an answer if it matches the
formula's answer within the benchmark's tolerances (the same status, a cushion within 0.5 points, a capacity within
0.5\% of current debt, and an exact match for ``none'' or ``undetermined''), and we flag it otherwise, including every
question the formula cannot answer. The formulas catch 98.0 to 99.7\% of the errors of open models up to ten times
larger, and 95.6\% of the errors of a proprietary frontier model whose remaining errors are the hardest to catch because its
accuracy is already high. The accepted answers are 99.0 to 99.9\% correct, whatever the quality of the LLM. The
untuned Mistral-7B reader answers only 24.1\% of questions correctly, yet the answers that the verifier accepts from
it are 99.0\% correct. On the clauses where two formalizers provably agree (92\% of clauses, covering 89\% of
questions), false flags fall from about 5\% to at most 0.8\%, while 94.4 to 99.7\% of errors are still caught.

\begin{table}[t]
\centering
\footnotesize
\definecolor{oursbg}{HTML}{E6F2E8}
\setlength{\tabcolsep}{3pt}
\setlength{\aboverulesep}{0pt}\setlength{\belowrulesep}{0pt}
\renewcommand{\arraystretch}{1.2}
\caption{FormaFin checking other LLMs' answers (\%, mean of three seeds, 95\% intervals over covenants). Although the
LLMs range from 24.1 to 97.0\% accuracy, the answers FormaFin accepts are 99.0 to 99.9\% correct. \emph{Accepted
correct} is the share of accepted answers that are correct, \emph{errors caught} the share of wrong answers flagged,
and \emph{false flags} the share of correct answers flagged. Rates are pooled over all 2,527 questions.}
\label{tab:llmverify}
\begin{tabular}{lc>{\columncolor{oursbg}}ccc}
\toprule
 & & \multicolumn{3}{c}{Checked by FormaFin} \\
\cmidrule(l){3-5}
LLM & Accuracy & Accepted correct & Errors caught & False flags \\
\midrule
\llmicon{mistral}Mistral-7B & 24.1 & 99.0\,{\tiny\textcolor{gray}{[97.8, 99.9]}} & 99.7\,{\tiny\textcolor{gray}{[99.3, 100.0]}} & 5.3\,{\tiny\textcolor{gray}{[1.4, 10.0]}} \\
\llmicon{meta}Llama-3.1-8B & 50.1 & 99.3\,{\tiny\textcolor{gray}{[98.6, 99.8]}} & 99.3\,{\tiny\textcolor{gray}{[98.7, 99.8]}} & 4.3\,{\tiny\textcolor{gray}{[1.7, 7.5]}} \\
\llmicon{qwen}Qwen2.5-Coder-7B & 61.6 & 99.2\,{\tiny\textcolor{gray}{[98.2, 99.9]}} & 98.7\,{\tiny\textcolor{gray}{[97.3, 99.9]}} & 4.9\,{\tiny\textcolor{gray}{[2.0, 8.1]}} \\
\llmicon{meta}Llama-3.3-70B & 78.0 & 99.7\,{\tiny\textcolor{gray}{[99.4, 99.9]}} & 99.0\,{\tiny\textcolor{gray}{[98.0, 99.8]}} & 5.5\,{\tiny\textcolor{gray}{[2.3, 9.2]}} \\
\llmicon{qwen}Qwen2.5-Coder-32B & 83.6 & 99.6\,{\tiny\textcolor{gray}{[98.9, 100.0]}} & 98.0\,{\tiny\textcolor{gray}{[94.9, 100.0]}} & 4.9\,{\tiny\textcolor{gray}{[1.9, 8.4]}} \\
\llmicon{qwen}Qwen3.8-27B & 84.8 & 99.8\,{\tiny\textcolor{gray}{[99.5, 100.0]}} & 99.1\,{\tiny\textcolor{gray}{[97.4, 100.0]}} & 4.5\,{\tiny\textcolor{gray}{[1.9, 7.7]}} \\
\llmicon{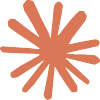}Claude Sonnet 5.5 & 97.0 & 99.9\,{\tiny\textcolor{gray}{[99.6, 100.0]}} & 95.6\,{\tiny\textcolor{gray}{[87.5, 100.0]}} & 5.7\,{\tiny\textcolor{gray}{[2.5, 9.3]}} \\
\bottomrule
\end{tabular}
\end{table}

\subsection{RQ4: Trusting formulas without labels}
\label{sec:verification}

\begin{figure}[t]
\centering
\includegraphics[width=0.85\textwidth]{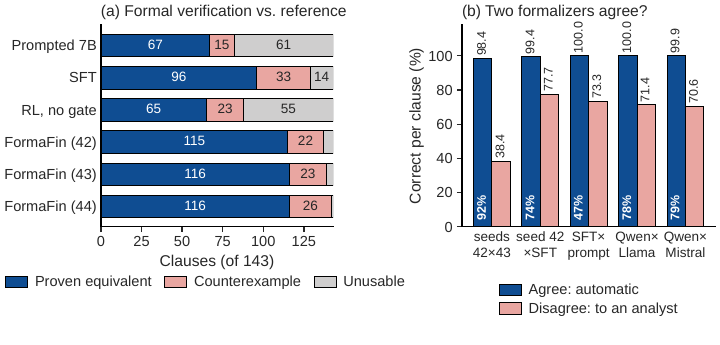}
\caption{Which formulas to trust. (a)~Z3 checks each formula against the reference. It proves the two equivalent,
finds a counterexample, or cannot use the formula. \emph{RL, no gate} is an earlier Stage II variant that accepts any parameter list and leaves the reporting date unconstrained during reward tests. (b)~Two formalizers are compared with no reference. The white
number is the share of covenants where they provably agree. On those covenants the answers are 98--100\% correct,
and the rest go to an analyst.}
\label{fig:verification}
\end{figure}
We first ask how often a formula provably matches the reference reading. On average over three seeds, Z3 proves 115.7 of the 143 FormaFin formulas equivalent to the reference, so they agree on every certificate in the defined input domain, not only those we constructed (\cref{fig:verification}a).

The \emph{RL, no gate} run illustrates why the parameter-list check matters. Unlike FormaFin, it trained on all 1,000 Stage II clauses rather than the 828 filtered clauses. Its reward accepted any parameter list and did not constrain the reporting date during limit tests. As a result, 48 of its 143 formulas omit the reporting date, and 7 more fail to parse. Accuracy falls to 58.4\% under the required interface, but reaches 91.3\% when missing parameters are accepted, above Stage I's 90.0\% under the same evaluation.

The reference exists only for evaluation. In deployment, we instead compare the formulas of two separately trained
seeds. They provably agree on 132 of 143 clauses (92\%), where the formalizer scores 98.4\%, and disagree on 11, where it scores only 38.4\%, so disagreement singles out the covenants an analyst should review (\cref{fig:verification}b). Agreement is not proof of correctness, because both seeds start from the same Stage I model and can make the same mistake. Z3 proves 113 of the 132 agreed formulas equivalent to the reference. The other 19 disagree with it on some certificates, 16 of them only at reporting date on or before the day a limit takes effect. Pairing models from different families, which share only the teacher's data, agrees on fewer covenants (78--79\%) but is right on 99.9--100\% of them, so a lender can choose its own balance between automation and accuracy.
\textbf{Proof versus testing.} Across three seeds, 397 formulas answer every benchmark certificate correctly, but Z3
proves that 50 of them (12.6\%) differ from the reference, 47 by applying a covenant before its effective date, which
no benchmark certificate covers. We also test the 71 formulas proven different on 1,000 random certificates each,
with amounts uniform on a log scale from \$1M to \$1B, dates from 1990 to 2030, and each condition true with
probability 0.5. This still misses two differences. A proof covers the whole domain and returns a certificate for
any difference. The advantage comes from proving rather than from SMT-LIB.

\section{Limitations}
\label{sec:discussion}
We construct the certificate figures, so the benchmark tests the computation on real clause
text, not real compliance rates, and it gives each defined term as one amount. Our references are two agreeing model
readings validated by the solver, not expert annotations; keeping only agreed clauses, and only clauses for which certificates can be constructed, favors simpler covenants, and
the teacher also produced one of the two reference specifications. Clauses with date schedules or conditions remain harder: FormaFin scores
$88.8\pm2.4\%$ on these 27 clauses against $96.3\pm0.9\%$ on the others. We ran the Llama-3.3-70B reader in 4-bit, which may
cost it some accuracy, though far less than its 15.7-point deficit. Our evaluation is limited to the borrowers of US public companies with agreements mostly
from 1996 to 2008. A proprietary teacher reads the
training clauses for Stages I and II. These clauses come from public SEC filings, and deployment calls no teacher, but
an institution that trains on its own confidential rules would need an open teacher, which we do not test.

\section{Conclusion and Future Work}
We introduced FormaFin, a framework for training small open models to translate financial rules into reusable SMT formulas. Our results show that separating rule interpretation from computation can improve the reliability of LLM-based financial analysis. On \bench{}, a 7B formalizer achieves 94.8\% accuracy, outperforming models ten times larger, while its formulas detect 95.6--99.7\% of errors in other LLMs' analyses. Solver-based rewards enable training without reference formulas during reinforcement learning, while equivalence proofs help identify formulas for automated use, although agreement alone does not establish correctness. Extending FormaFin to other financial rules, including regulatory ratios and investment mandates, remains future work.

\bibliographystyle{splncs04}
\bibliography{references}

\end{document}